\documentclass[letterpaper, 10 pt, conference, twocolumn]{ieeeconf}
\IEEEoverridecommandlockouts
\usepackage[utf8]{inputenc}

\usepackage{graphicx}
\usepackage[nolist]{acronym}
\usepackage{amsmath}
\usepackage{amssymb}

\usepackage{hyperref}
\hypersetup{
    bookmarks=false,
    colorlinks=true,
    linkcolor=blue,
    filecolor=magenta,      
    urlcolor=blue,
    citecolor=black,
}

\usepackage[capitalise]{cleveref}
\usepackage{float}
\usepackage[caption=false,font=footnotesize]{subfig}

\newcommand{\Tau}{\mathrm{T}}

\begin{acronym}
\acro{ACP}{Attention-based Costmap Prediction}
\acro{AIRL}{Approximate Inverse Reinforcement Learning}
\acro{BNN}{Bayesian Neural Network}
\acro{CNN}{Convolutional Neural Network}
\acro{DNN}{Deep Neural Network}
\acro{DAgger}{Data Aggregation}
\acro{DDP}{Differential Dynamic Programming}
\acro{DL}{Deep Learning}
\acro{E2E}{end-to-end}
\acro{IL}{Imitation Learning}
\acro{iLQG/MPC-DDP}{iterative Linear Quadratic Gaussian/Model Predictive Control Differential Dynamic Programming}
\acro{IRL}{Inverse Reinforcement Learning}
\acro{LSTM}{Long Short Term Memory}
\acro{Conv-LSTM} {Convolutional LSTM}
\acro{MDP}{Markov Decision Process}
\acro{MPC}{Model Predictive Control}
\acro{MPPI}{Model Predictive Path Integral control}
\acro{NN}{Neural Network}
\acro{RL}{Reinforcement Learning}

\acroplural{MDP}[MDPs]{Markov Decision Processes}
\end{acronym}

\title{\LARGE \bf
\acl{AIRL} from Vision-based Imitation Learning
}

\author{Keuntaek Lee, Bogdan Vlahov, Jason Gibson, James M. Rehg, and Evangelos A. Theodorou% <-this % stops a space
\thanks{*This work was supported by Amazon Web Services (AWS) and NASA Langley Research Center Grant 80NSSC19M0211.}% <-this % stops a space
\thanks{The authors are with the Georgia Institute of Technology, Atlanta, GA, USA.
        Correspondence to: {\tt\small keuntaek.lee@gatech.edu}}%
}

\usepackage{fancyhdr}
\begin{document}

\maketitle
\thispagestyle{fancy}
\pagestyle{empty}

%%%%%%%%%%%%%%%%%%%%%%%%%%%%%%%%%%%%%%%%%%%%%%%%%%%%%%%%%%%%%%%%%%%%%%%%%%%%%%%%
\begin{abstract}

In this work, we present a method for obtaining an implicit objective function for vision-based navigation. The proposed methodology relies on Imitation Learning, Model Predictive Control (MPC), and an interpretation technique used in Deep Neural Networks. We use Imitation Learning as a means to do Inverse Reinforcement Learning in order to create an approximate cost function generator for a visual navigation challenge.  The resulting cost function, the costmap, is used in conjunction with MPC for real-time control and outperforms other state-of-the-art costmap generators in novel environments. The proposed process allows for simple training and robustness to out-of-sample data. We apply our method to the task of vision-based autonomous driving in multiple real and simulated environments and show its generalizability.
\\Supplementary video:
\textcolor{blue}{https://youtu.be/WyJfT5lc0aQ}
\end{abstract}

%%%%%%%%%%%%%%%%%%%%%%%%%%%%%%%%%%%%%%%%%%%%%%%%%%%%%%%%%%%%%%%%%%%%%%%%%%%%%%%%
\section{INTRODUCTION}

\acresetall

In robotics, vision-based control has become a popular topic as it allows navigating in a variety of environments.
A notable contribution is the ability to work in areas where positional information from satellites or motion capture systems is not possible to obtain.
While vision-based controls are harder to analytically write equations for when compared to positional-based controls, it has been shown to work by millions of humans using it every day.

Using deep neural networks (NNs) for vision-based control has become ubiquitous in literature thanks to the power of deep learning.
% In the recent autonomous driving literature, \cite{drews2019visual} used an architecture that separates the vision-based control problem into a cost function generation task and then used an MPC controller for path planning and optimal control. This method allows for the less principled area of feature extraction and interpretation for autonomous driving to be done by the NN, and solve the stochastic optimal control problem in a principled way. This architecture provided better observability into the learning process as compared to traditional \ac{E2E} control approaches \cite{bojarski2016end, xu2017end, Pan2018RSS}. Additionally, this decouples the state estimation and controller, allowing us to leverage standard state estimation techniques with a vision-based controller.
However, in general, most of the NN models suffer from the generalization problem or the covariate shift; a trained NN model does not work well on a new test dataset if the distribution of input variables from training and testing dataset are very different from each other.

To solve this generalization problem in new environments, in this work, we focus on generalizing vision-based control systems to new previously unseen environments. This will focus on the ability of a single network to generate reasonable cost functions even in a novel environment not seen during training.
We propose an automatic way to generate the grid map cost function, the costmap, without requiring access to a pre-defined costmap or any labels with which to perform segmentation, classification, or recognition. The key idea is using a vision-based \ac{E2E} \ac{IL} framework \cite{Pan2018RSS}. In this \ac{E2E} control approach, we only need to query the expert's action to learn a costmap of a specific task. During training, the model learns a mapping from sensor input to the control output. Specifically in vision-based autonomous driving, if we train a \ac{DNN} by imitation learning and analyze an intermediate layer of \acp{CNN} by reading the activated neurons of the trained network, we see the mapping converged to extracting important features that link the input and the output.

In a broad sense, the convolutional layer parts of the trained \ac{E2E} network become a function that extracts important features in the input scene.
This can be viewed as an implicit image segmentation done inside the deep \ac{CNN} where the extracted features will depend on the task at hand. For example, if the task is learning to visually track an object, the network will implicitly find the object as an important feature. In another case, if the task is to perform autonomous lane-keeping, the boundaries of the lane will become important for making a final decision.

Our work is obtaining a costmap based on an intermediate convolutional layer activation, but the middle layer output is not directly trained to predict a costmap; instead, it is generating an implicit objective function related to relevant features, which links the input and the output. This allows our work to produce a reasonable costmap on unseen data where direct costmap prediction methods \cite{drews2019visual} would fail because the data would be out of their prediction domain.
% As an analogy, our method is similar to learning the addition operator $a+b = c$ whereas a prediction method would be similar to a mapping between numbers $(a,b) \rightarrow c$, which fails at test time if we never saw the same numbers $(a,b)$ during learning, i.e. if $(a,b)$ is not in the lookup table.

% \section{RELATED WORK}
Monocular vision-based planning has shown a lot of success performing visual servoing in ground vehicles \cite{Drews2017corl, Drews2019RAL, drews2019visual}, manipulators \cite{LevineVisuomotor}, and aerial vehicles \cite{giusti2016machine, lee2020pixelmpc}.
In the autonomous driving literature, \cite{Drews2019RAL} learned to generate a costmap from camera images for the \ac{MPC} controller. \cite{drews2019visual} tried to generalize this approach by using a \ac{Conv-LSTM} and a softmax attention mechanism and shows this method working on previously unseen tracks. However, the training of this architecture requires having a predetermined costmap to imitate and the track it was shown to generalize had visually similar components (dirt track and black track borders) to recognize.

% \cite{loquercio2019deep} constructs a system for vision-based agile drone flight that generalizes to new environments. They separate their system into a perception and control pipeline. The perception pipeline was a \ac{CNN}, taking in raw images and producing a desired direction and velocity, trained in simulation on a large mixture of random backgrounds and gates. By providing many random textures, the perception pipeline is trained to be more generalized than training on just real data. They show this system can perform similarly or better than a system trained on real-world data alone from real drones. However, their method is still best applied to drones where it is relatively easy to match a desired direction and velocity.

End-to-end learning in autonomous driving has been shown to work in various lane-keeping applications  \cite{bojarski2016end, xu2017end} and \cite{Pan2018RSS} showed great performance by learning both steering and throttle but did not show its generalizability except for different lighting conditions.

\cite{Ollis2008outdoor} proposed an image-space approach for vision-based navigation. The approach plans a path for a ground-based robot in the image-space of an onboard monocular camera. This technique is most related to our approach since they applied a learned color-to-cost mapping to transform a raw image into a costmap-like image, and performed path planning directly in the image space.
The limitation of \cite{Ollis2008outdoor} is that they require another supervised learning step to train a model to output the costmap like in \cite{drews2019visual}, whereas our method does not need one. Therefore, our approach is more efficient in learning a costmap without having a prior on what the costmap should look like.

% \subsection{Contribution}
The contributions of this work are threefold: 
% \begin{inparaenum}[i)]
\begin{enumerate}
  \item We introduce a novel inverse reinforcement learning method \ac{AIRL} which approximates a cost function from an intermediate layer of an end-to-end policy trained with imitation learning.
  \item We perform a sampling-based stochastic optimal control, MPPI, in image space, which is perfectly suitable for our driver-view binary costmap.
  \item Compared to state-of-the-art methods, the \ac{ACP} and the \acl{E2E} \acl{IL} (\ac{E2E}\ac{IL}), our proposed method is shown to generalize well by generating usable costmaps in environments outside of its training data.
\end{enumerate}
% \end{inparaenum}

% The remaining of the paper is organized as follows: In \Cref{sec:preliminaries}, we briefly review some preliminaries used in our work with some literature reviews. \Cref{sec:mppi_imagespace} introduces the \ac{MPPI} control algorithm in image space and in \Cref{subsec:AIRL}, we introduce our \acl{AIRL} algorithm. \Cref{sec:experiments} details vision-based autonomous driving experiments with analysis and comparisons of the proposed methods. Finally, we conclude and discuss future directions in \Cref{sec:discussion} and \Cref{sec:conclusion}.

\section{BACKGROUND}
\label{sec:preliminaries}

\subsection{\acl{IRL}}

In \ac{RL}, given that an agent is able to query the reward of applying any action at any state, the goal is to find the optimal policy that maximizes the expected future reward.
In \ac{IRL}, the underlying reward function is unknown and the goal is to find the reward function that explains the given optimal policy.
\ac{IRL} can be considered a harder problem to solve than \ac{RL}, because generally, there is no single reward function that can describe an expert behavior \cite{arora2018survey}.

If we can approximate a reward function from observations, we can then train new agents to maximize this reward. It can be considered similar to \acl{IL} in that sense, as we could train agents to perform according to an expert behavior. However, it is important to note that, unlike in \ac{IL}, the learning agents could then potentially outperform the expert behavior. This is especially true when the expert behavior is suboptimal or applied in a different environment.
% \cite{Russell1998} and \cite{arora2018survey} also describes how a learned reward function is more transferable than an expert policy because as a policy can be easily affected by different transition functions $T$ whereas the reward function can be considered a description of the ideal policy.

Most of the \ac{IRL} work in the literature requires one more pipeline of training to figure out the mapping between the input trajectories and the reward function. For example, \cite{Babes2011IRL} introduced maximum likelihood approach to \ac{IRL} while the maximum entropy approach was introduced in \cite{Kuderer2015ICRA} to find a generative model that yields trajectories that are similar to the expert's. They find a weighted distribution of reward basis functions in an iterative way. This step still requires some hand-tuning; for example, picking proper basis functions to form the distribution. 

However, in our work, we do not need any new design of basis functions because we learn a policy \acl{E2E} and can get an approximate the reward function for `free', i.e. without any hand-tuning. Since optimal controllers can be considered as a form of model-based \ac{RL}, the negative of this reward function can then be used as the cost function that our \ac{MPC} controller optimizes with respect to.

\subsection{\acl{IL}}

% \ac{RL} is one way to train agents to maximize some notion of task-specific rewards. One of the major problems in \ac{RL} is the sample-inefficiency problem: the agents have to explore the action-state space without any prior knowledge of the environment or task. However, \ac{IL} uses supervised learning to train a control policy and bypass this sample-inefficiency problem. 
In \acp{IL} (\ac{IL}), a policy is trained to accomplish a specific task by mimicking an expert's control policy, which in most cases, is assumed to be optimal.
% Accordingly, \ac{IL} provides a safer training process. 
In this work of perceptual control, we will use sections of a network trained with \acl{E2E} \acl{IL} (\ac{E2E}\ac{IL}) using \ac{MPC} as the expert policy. Literally, \ac{E2E}\ac{IL} trains agents to directly output optimal control actions given image data from cameras; end (sensor reading) to end (control).

While \ac{IL} provides benefits in terms of sample efficiency, it does have drawbacks. Here, we shortly talk about three major problems in \ac{IL}.

\textbf{i) Generalizability}:
% The training data collected from an optimal expert does not usually include demonstrations of failure cases in unsafe situations,
Even with online data aggregation methods \cite{DAgger}, it is impossible to collect all kinds of unexpected scenarios and edge cases.
% \cite{DAgger} introduced an online \ac{DAgger} method, which mixes the expert's policy and the learner's policy to explore various situations like $\epsilon$-greedy algorithm.
% However, even with the online scheme of collecting datasets,
% it is impossible to experience all kinds of unexpected scenarios.
Accordingly, \ac{E2E}\ac{IL} is vulnerable to out-of-training-data (covariate shift) as shown in the literature \cite{lee2019earlyfailure, lee2019ensemble, lee2019papc}. There are little to no guarantees on what a \ac{NN} trained with \ac{IL} will output when given an input vastly different from its training set.
% \cite{lee2019earlyfailure, lee2019papc, lee2019ensemble} demonstrated failure cases of deep end-to-end controllers; the controllers failed to predict a correct label from a novel (out-of-training-data) input and there was no way to tell the output prediction is trustworthy without considering the Bayesian technique.

\textbf{ii) Upper-bounded}:
The best job a learner can do is capped by the ability of a teacher since the objective of the \ac{IL} setting is to mimic the expert's behavior.

\textbf{iii) Interpretability}:
Since the end-to-end approach uses a totally blackbox model from sensor input to control output, it loses interpretability; when it fails, it is hard to tell if it comes from noise in the input, if the input is different from the training data, or if the model has just chosen a wrong control output due to ending training prematurely.

From these reasons, \ac{E2E} \ac{IL} controllers are not widely used in real-world applications, such as self-driving cars.
Our approach provides solutions to these problems by leveraging the idea of using \ac{DL} only in some blocks of autonomy, hence becomes more interpretable.

In the case of autonomous driving, given a cost function to optimize and a vehicle dynamics model, we can do path planning and compute an optimal solution via an optimal model predictive controller.
Therefore, the problem simplifies from computing a good action to computing a good approximation of the cost function in new environments.

In this paper, we provide evidence of better performance than the expert teacher by showing a higher success rate of task completion when a task requires generalization to new environments.

\section{MPPI IN IMAGE SPACE}
\label{sec:mppi_imagespace}

We used a stochastic MPC optimal controller, \ac{MPPI} \cite{mppi17}, as an expert in \ac{IL} and also as an optimal controller for testing our costmap, like in \cite{Drews2017corl, Drews2019RAL, drews2019visual}.
The details of the \ac{MPPI} algorithm can be found in multiple literature \cite{mppi16, mppi17} and here we concisely describe \ac{MPPI}; \ac{MPPI} is a sampling-based stochastic optimal controller which can operate on nonlinear dynamics and can have non-convex cost functions. It is an iterative optimization algorithm for path planning and control with a receding time horizon MPC.
% In this section, due to the lack of space, we focus more on describing our contribution part of the extended version of \ac{MPPI} \cite{mppi17}.

For the navigation task, our cost function for the optimal control problem will follow a similar format as in \cite{drews2019visual} with the squared cost on the desired speed. The cost function at time $t$, which does not penalize the control, is shown as:
\begin{align}
    l(\textbf{x}_t) = C_{speed}(v_{x}^{d}-v_{x,t})^2 + C_{crach} I(\textbf{x}_t),
    \label{eq:cost}
\end{align}
where $C_{speed}, C_{crash}$ are coefficients that represent the penalty applied for speed and crash, respectively. $I$ is an indicator function that returns $1$ if the vehicle position in the image space is on the high-cost region, and returns $0$ otherwise. $v_x$ and $v_x^d$ are measured body velocity in the $x$ direction and desired velocity respectively.
It is important to note that in our settings, near-perfect state estimation and a GPS-track-map is provided when \ac{MPPI} is used as the expert during data collection, but as in \cite{drews2019visual}, only body velocity, roll, and yaw from the state estimate is used when it is operating using vision at testing. For our navigation task, we followed the same definition of the system state and control in the MPPI paper \cite{mppi17} and \cite{drews2019visual}:  $\mathbf{x}=[x,y,yaw,roll,v_x,v_y,\dot{yaw}]$ is the vehicle state in a world coordinate frame and $\mathbf{u}$ is $[throttle, steering]$.

Image space from a mounted camera on a robot is a local and fixed frame; i.e. the state represented in the image space is relative to the robot's camera.
Since we are planning an optimal path given a costmap image in first-person-view, the vehicle's future state trajectory described in the world coordinates must be transformed into a 2D image in a moving frame of reference. This traditional coordinate transformation technique is widely used in 3D computer graphics \cite{coordtransformbook}. %and described in the Appendix \Cref{sec:appendix_coord_trans}.

\begin{figure}[H]
  \centering
  \vspace{-0.4cm}
    \subfloat[The  original  output  from  the  deep  CNN  middle  layer. The  white  colored  pixels  are  the  activated  neurons  averaged  among  the convolution filters.  \label{fig:costmap_original}]{\includegraphics[width=0.15\textwidth]{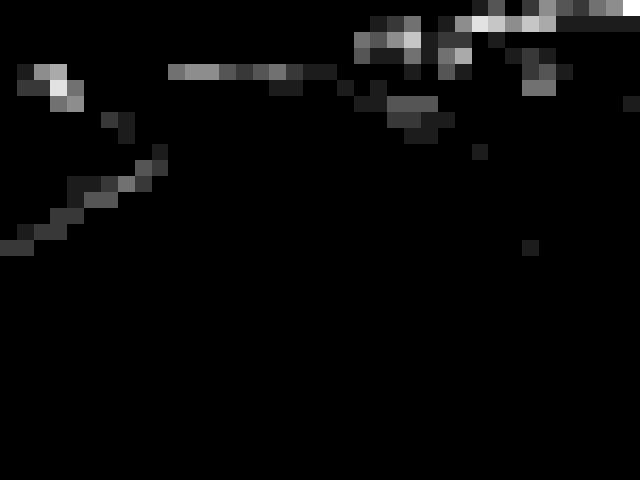}}
    \vspace{0.5mm}
    \subfloat[After applying a binary filter. The black and white background represents the costmap that MPPI is optimizing. \label{fig:costmap_gaussian_threshold}]{\includegraphics[width=0.15\textwidth]{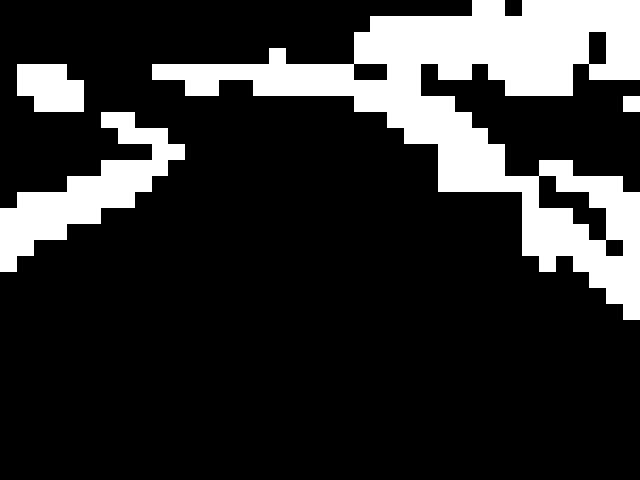}}
    \vspace{0.5mm}
    \subfloat[MPPI in image space. The green trajectory is the MPPI-planned future state trajectory in image space. \label{fig:mppi_imagespace}]{\includegraphics[width=0.15\textwidth]{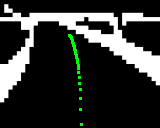}}
  \caption{Running MPPI on image space with the generated costmap.\label{fig:costmaps}}
  \vspace{-0.1cm}
\end{figure}

Through the coordinate transform at every timestep, the MPPI-planned final future state trajectory mapped in image space on our costmap looks like \Cref{fig:mppi_imagespace}.

% We use the total cost function \Cref{eq:costfunction} with a running cost \Cref{eq:cost} and zero terminal cost for an autonomous driving task. In this formulation of MPPI, two cost terms predominate: speed and track-related crash costs.
The crash cost depends on the costmap and it is a binary grid map (0, 1) describing occupancy of features we want to avoid driving through, e.g. track boundaries or lane boundaries on the road.
% Going off the image plane does not have a cost associated with it.

\section{METHODOLOGIES}
\label{sec:methods}
In this work, we introduce a method for an inverse reinforcement learning problem with the task of vision-based autonomous driving.
More specifically, we focus on lane-keeping and collision checking like in \cite{Drews2017corl, Drews2019RAL, drews2019visual, Pan2018RSS, bojarski2016end}. The following methods are evaluated in \Cref{sec:experiments}.
\subsection{End-to-End \acl{IL} \cite{Pan2018RSS}}
\label{sec:Yunpeng}
\cite{Pan2018RSS} constructed a \ac{CNN} that takes in RGB images and spits out control actions of throttle and steering angles for an autonomous vehicle. It was trained using \ac{DAgger} algorithm \cite{DAgger} on data provided by an expert \ac{MPC} controller. While it can achieve aggressive driving targets and was shown to handle various lighting conditions on the same track, it in general does not generalize to brand new tracks. This is most likely due to the images creating a feature space not seen in training. While the last layers may not be able to choose the proper action, intermediate layers still perform some feature extraction. This feature extraction is further discussed in \Cref{subsec:AIRL}.
\subsection{\acl{ACP} \cite{drews2019visual}}
\label{sec:Paul}

The concise description of \cite{drews2019visual}, we call \acl{ACP} (\ac{ACP}), is to create a \ac{NN} that can take in camera images and output a costmap used by a \ac{MPC} controller, \ac{MPPI}. 
% The major contributions over \cite{Drews2017corl, Drews2019RAL} are using a \ac{Conv-LSTM} layer to maintain the spatial information of states close together in time as well as a softmax attention mechanism applied to sparsify the \ac{Conv-LSTM} layer. They show that the attention image mimics areas where humans focus when driving a vehicle, which provides evidence of a generalization technique similar to humans.  The costmap generated by this \ac{NN} architecture is then provided to an \ac{MPPI} controller. 
By separating the perception and low-level control into two robust components, this system can be more resilient to small errors in either. Their final model is trained on a mixture of real datasets of a simple racetrack as well as simulation datasets from a more complex track. The full system is then able to drive around the real-world version of the complex track in an aggressive fashion without crashing. It is the most generalized method for achieving autonomous driving in new environments that the authors of this paper have found in the literature.

\subsection{\acl{AIRL} (ours)}
\label{subsec:AIRL}
Our method can be considered a mixture of the two previously mentioned; we will be using both \ac{E2E}\ac{IL} and an \ac{MPC} controller.
Although our work relies on \ac{E2E}\ac{IL} and \ac{MPC}, we tackle a totally different problem: IRL from \ac{E2E}{IL}. Our main contribution is learning an approximate, `generalizable' costmap `from' \ac{E2E}\ac{IL}. On top of this \ac{AIRL}, we perform \ac{MPC} in image space (\Cref{sec:mppi_imagespace}) with a real-time-generated agent-view costmap.
The reason why we named it `Approximate' IRL is because we do not directly train a model to output a cost function.

% To repeat our problem statement, it is an inverse reinforcement learning problem of learning a cost function and the task is autonomous driving.

To support our method, we borrow the ideas from one of the interpretation techniques used for interpreting the information flow inside \acp{DNN}. Pixel-wise heatmaps or activation maps have been widely used to interpret and explain the deep CNN’s predictions and the information flow, given an input image \cite{interpretingDNN, explainableAI, samek2017evalvis}. The Layer-wise Relevance Propagation (LRP) \cite{lrp} is one of the state-of-the-art methods developed to understand, visualize, interpret, and explain a `trained' nonlinear black-box models (e.g. \acp{DNN}) and decisions made by them in machine learning problems. The explanation process relies on a \textit{heatmap} visualizing each pixel's contribution and the relevance to the prediction. LRP is invariant against the choice of nonlinear activation functions and also max-pooling fits into the structure. LRP follows the layer-wise relevance conservation rule:
\begin{align}
\sum_i \mathcal{R}_i=...=\sum_j \mathcal{R}_j=\sum_k \mathcal{R}_k=...=f(x), \label{eq:relconserv}
\end{align}
where $\mathcal{R}_i$ is the relevance score at layer $i$, $x$ is the input, $f$ is the decision model, and $f(x)$ outputs the decision. This rule says that at every layer of a DNN, the total relevance is conserved. The redistribution rule is defined \cite{lrp} as
\begin{align}
\mathcal{R}_j=\sum_k\frac{x_j w_{jk}}{\sum_h x_h w_{hk}} \mathcal{R}_k. \label{eq:backward_redist}
\end{align}
This definition basically redistributes relevance from layer $l+1(=k)$ to $l(=j)$ in a backward way, starting from the output $f(x)$. The redistribution is done proportional to the neuron activation $x_j$ at layer $l$, and the strength of the weights $w_{jk}$, i.e. the larger the activation $x_j$ is and the larger the connection $w_{jk}$ is, the more relevance flows through them. The denominator term is a sum of all $x_h w_{hk}$ between layer $k$ and its previous layer, and works as a normalizing constant.

Here, we introduce a forward redistribution rule, based on the layer-wise relevance conservation rule, \Cref{eq:relconserv}: \begin{align} \mathcal{R}_k=\sum_j\frac{x_j w_{jk}}{\sum_h x_h w_{hk}} \mathcal{R}_j. \label{eq:forward_redist} \end{align} \\ This new rule talks about the relevance scores propagating from the input to the output. Both backward and forward rules, \Cref{eq:backward_redist} and (\ref{eq:forward_redist}), basically explain the same thing but in the opposite ways. \textbf{Backward:} the final output is more affected by more highlighted region of previous layer and stronger connections between them. \textbf{Forward:} more highlighted region of a layer and stronger connections between the layer and the output more leads to the results.

If we name the heatmap coming from the relevance scores in the backward redistribution rule, the \textit{backward heatmap}, then the heatmap coming from the relevance scores in the forward rule can be named as the \textit{forward heatmap}. Note that the \textit{forward heatmaps} are nothing more than the normalized neuron activations from the forward propagation in DNNs.

Although the backward redistribution is initialized with the output $f(x)$, the forward redistribution does not need to be initialized. After the input image is passed through the first layer of DNN, the normalized relevance scores will indicate which pixel of the input image has more relevance scores to the output.

% \textcolor{red}{TODO: explain backward and forward redistribution with figures}

The importance of the \textit{forward heatmap} is a) We can get the costmap just through the inference step of a trained DNN. b) From this fact, we only need to enact the first few parts of the DNN which is used to output the costmap. c) If we want to get the costmap from the \textit{backward heatmaps}, it requires the full model of a \ac{DNN} and it requires a round-trip, from input to the output, and again from output to a middle layer. Therefore, it is slower than the \textit{forward heatmap} and inefficient.

% To give an analogy in optimal control, we can think about the backward redistribution rule as the backward Bellman optimality principle, which explains the relationship between the optimal solution and the value function by backward induction $(t=T,...,0)$. Then the forward redistribution rule corresponds to applying the solution forward $(t=0,...,T)$ and obtaining the total value function at the end. 

To exploit the LRP rules in our imitation learning framework, we make one assumption here:\\ \textit{\textbf{Assumption 1}: If a DNN model converged in the imitation learning fashion, i.e., the learner mimics the expert almost perfectly, then the neurons in each layer also converged and have reasonable forward relevance scores.}

It has been empirically shown in \cite{lrp, samek2017evalvis} that the \textit{backward heatmaps} converged to `reasonable' values after training DNNs in a typical supervised-learning fashion. Here, the `reasonable' scores in our image-based CNN framework means that the scores must make sense to a human. For an example of driving the road, to get the decision of `turn left', in any middle layer, the region corresponding `left turn' sign on the input image should be highlighted, not other regions corresponding objects like trees or houses in the input.

\begin{figure}[H]
    \centering
    \subfloat[50]{\includegraphics[width=0.075\textwidth]{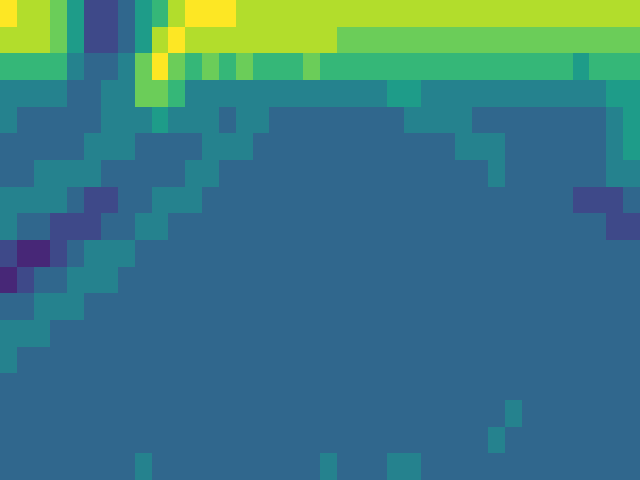}}
    \hspace{0.1mm}
    \subfloat[100]{\includegraphics[width=0.075\textwidth]{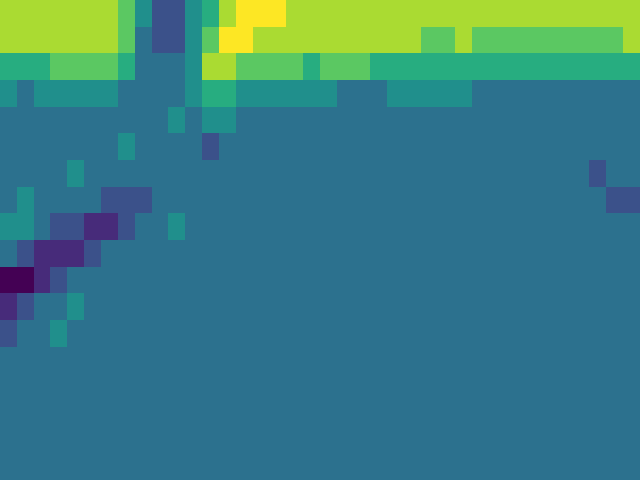}}
    \hspace{0.1mm}
    \subfloat[150]{\includegraphics[width=0.075\textwidth]{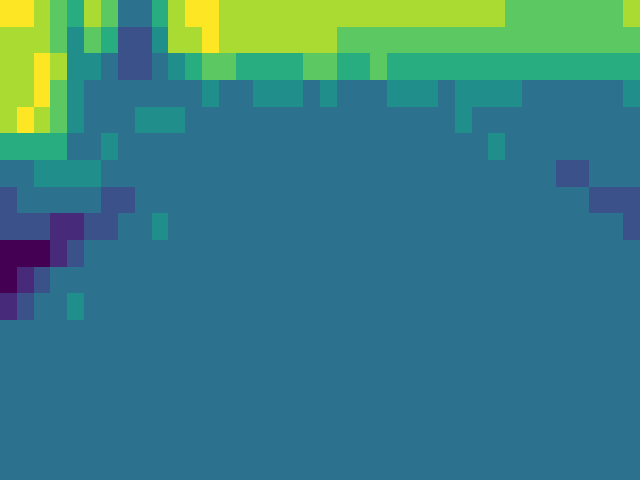}}
    \hspace{0.1mm}
    \subfloat[200]{\includegraphics[width=0.075\textwidth]{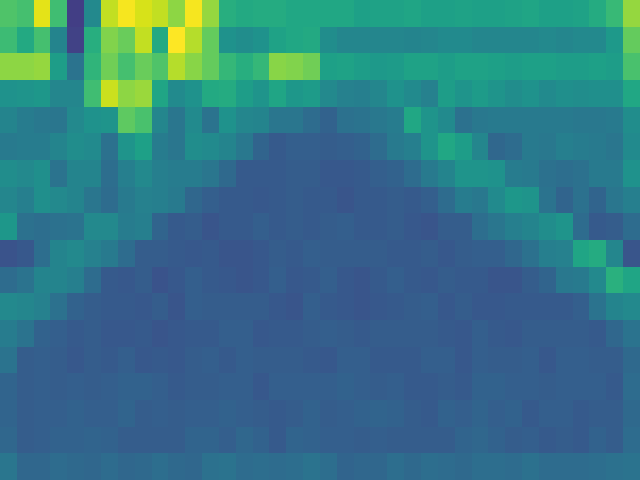}}
    \hspace{0.1mm}
    \subfloat[250]{\includegraphics[width=0.075\textwidth]{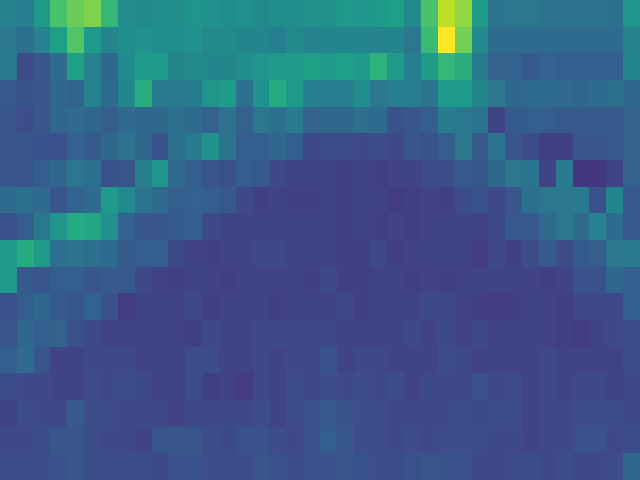}}
    \hspace{0.1mm}
    \subfloat[300]{\includegraphics[width=0.075\textwidth]{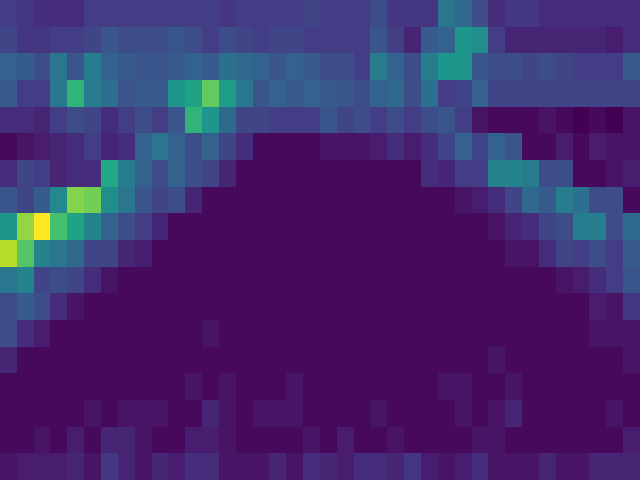}}
    \caption{Forward heatmap plots at every 50 training epochs of end-to-end imitation learning. Brighter color represents more activation of neurons. After 300 iterations of training, the lanes (track boundaries) are more highlighted than other features. These plots emperically show that \textit{\textbf{Assumption 1}} holds.}
    \label{fig:assumption1}
\end{figure}

LRP then allows us to extract any middle layer of DNNs and use the \textit{forward heatmap} to interpret and infer about the relationship between input, heatmap, and the output. We empirically show in \Cref{fig:assumption1} that \textit{\textbf{Assumption 1}} holds.

We then further interpret this intermediate stage, the \textit{forward heatmap}, as cost function-related important features that relate the input (observation) and the output (final optimal decision) under the optimal control settings.
The averaged heatmap extracted from a middle convolutional layer of the trained \ac{E2E}\ac{IL} network is used to generate a costmap for \ac{MPC}.

% Our proposed approach requires one more typical assumption:\\
% \textit{\textbf{Assumption 2}: Like in any classic \ac{IL} settings, the expert's behavior is optimal.}

% In this work, we used a model predictive optimal controller, \ac{MPPI} \cite{mppi17}, as the expert for \ac{E2E}\ac{IL}. This is similar to \cite{drews2019visual} in that we have separated the perception pipeline from the controls.

\begin{figure}
  \centering
  \includegraphics[width=0.49\textwidth]{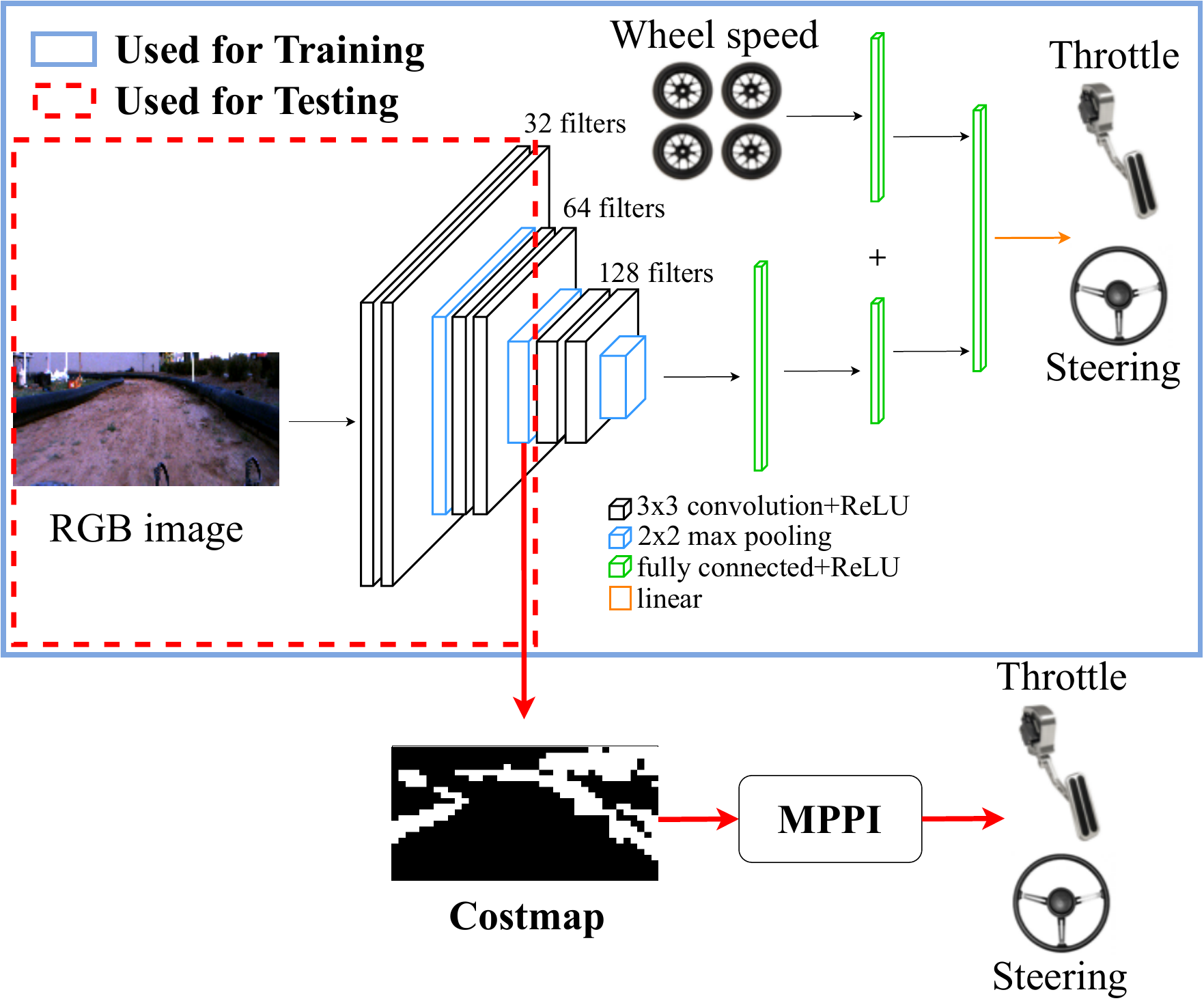}
  \vspace{-0.5cm}
  \caption{\textit{Blue box}: The same structure used in \cite{Pan2018RSS} for \ac{IL} training. For \ac{IRL} testing, only the red-dashed part, from the image input to the second max pooling layer is used. This middle layer internally and implicitly learns the mapping from an input image to important features, which can be used as a costmap.}
  \label{fig:structure}
  \vspace{-0.3cm}
\end{figure}

The training process is the same as the \ac{E2E}\ac{IL} controller \cite{Pan2018RSS}; \ac{AIRL} only requires a dataset of images, wheel speed sensor readings, and the expert's optimal solution to train a costmap model (see \Cref{fig:structure}). 
% The dimensions in detail can be found in Appendix \Cref{subsec:dim}.

Unlike \cite{Drews2017corl, Drews2019RAL, drews2019visual}, our method does not require access to a predetermined costmap function in order to train. Also, due to the fact that \ac{E2E}\ac{IL} can be taught from human data only \cite{bojarski2016end}, our approach can learn a cost function even without teaching specific task-related objectives to a model.

% We would like to use all of the activated middle layer neurons to generate a costmap, but the magnitude of the activation is different for each feature since they have different relevance scores.
% However, since we consider all the activated features important for a costmap, we add a binary ($0$ or $1$) filter. The binary filter outputs 1 if the activation is greater than 0, i.e. if it has some relevance scores.
We generate a costmap from the activated middle layer neurons, which links the input image and the cost function. On top of that, we add a binary ($0$ or $1$) filter which outputs 1 if the activation is greater than 0, i.e. if it has some relevance scores.
% In this way, we equally regard all the activated features as important ones. 
This binary filter is not a necessary step for an optimal controller, but rather heuristic to improve and robustify the performance. Even without the binary filter, the optimal controller eventually finds the optimal control policy that drives the robot to the drivable region but the binary filtering makes the costmap easier to optimize. The filtered costmap is more distinguishable between drivable region (0, black) and cost regions (1, white) as shown in \Cref{fig:costmaps}.
% Adding a binary filter may look like a simple step, but this is the biggest reason why our costmap generation is stable while the \ac{E2E} controller fails.
% \textcolor{red}{The binary filter links the extracted important features in the image to our optimal controller \Cref{sec:mppi_imagespace}, which optimizes a binary cost function. This cost function is composed of either 0 (drivable region) or 1 (obstacles).}

% We also introduce a risk-sensitive version of the \ac{AIRL} in Appendix \Cref{subsec:risksensitive}.
% In the next section, we show the experimental results of the vanilla \ac{AIRL} and leave some room for the risk-sensitive version for future works.

To describe the detailed dimensions used in \ac{AIRL}, the input image size is $160 \times 128 \times 3$ and the output costmap from the middle layer after the second max-pooling (See \Cref{fig:structure}) is $40 \times 32$.
We tested the heatmaps from 3 middle layers in the \ac{CNN} part, each of them were after each max-pooling layer. The 3 heatmaps have different sizes; $(80 \times 64)$, $(40 \times 32)$, and $(20 \times 16)$. The feature activation was more distinguishable in deeper layers (smaller heatmaps) as the information flows towards the output. But the resolution of the smallest heatmap was too low to be used as a costmap, so we decided to use the heatmap with size $(40 \times 32)$.
This 2D costmap comes from taking the average of the activated neurons with respect to all 128 kernels ($128 \times 40 \times 32 \times 3 \rightarrow 40 \times 32 \times 3$), and converting the 3D RGB channel into grayscale ($40 \times 32 \times 3 \rightarrow 40 \times 32$).
This is then resized to $160 \times 128$ for \ac{MPPI} costmap.

\section{EXPERIMENTS}
\label{sec:experiments}
%\subsection{Costmap prediction}

\subsection{Data collection}

\begin{figure}[b] % 0.15 0.28 0.22 0.19 0.22 0.19
    \centering
    \vspace{-0.7cm}
    \subfloat[The test vehicle \label{fig:car}]{\includegraphics[width=0.15\textwidth]{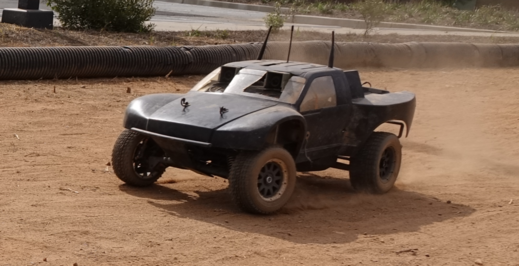}}
    \hspace{1mm}
    \subfloat[Track A \label{fig:CCRF}]{\includegraphics[width=0.15\textwidth]{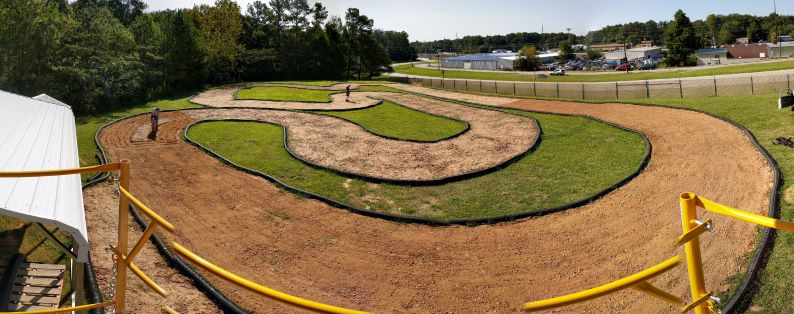}}
    \hspace{1mm}
    \subfloat[Track B \label{fig:Marietta}]{\includegraphics[width=0.15\textwidth]{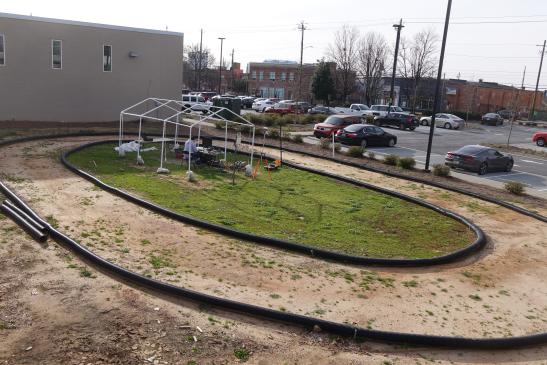}}
    \hspace{1mm}
    \subfloat[Track C \label{fig:tarmac}]{\includegraphics[width=0.15\textwidth]{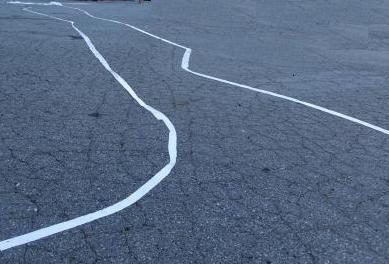}}
    \hspace{1mm}
    \subfloat[Track D \label{fig:Marietta_sim}]{\includegraphics[width=0.15\textwidth]{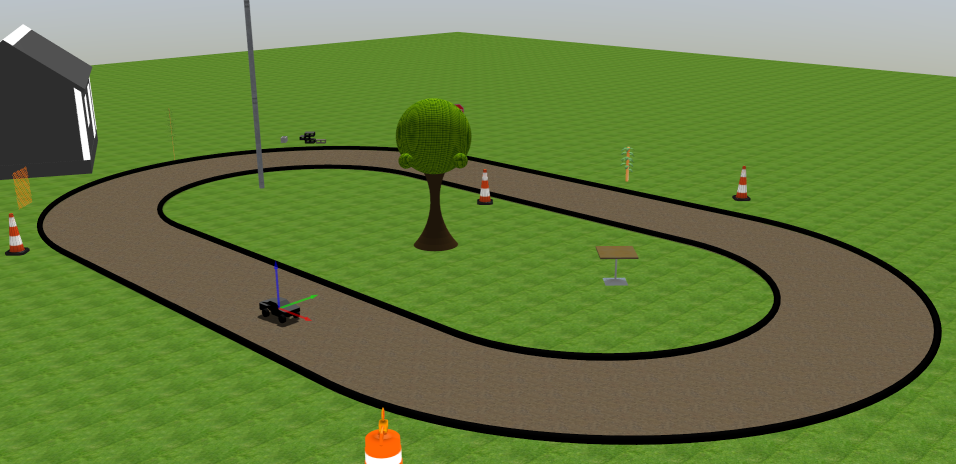}}
    \hspace{1mm}
    \subfloat[Track E \label{fig:CCRF_sim}]{\includegraphics[width=0.15\textwidth]{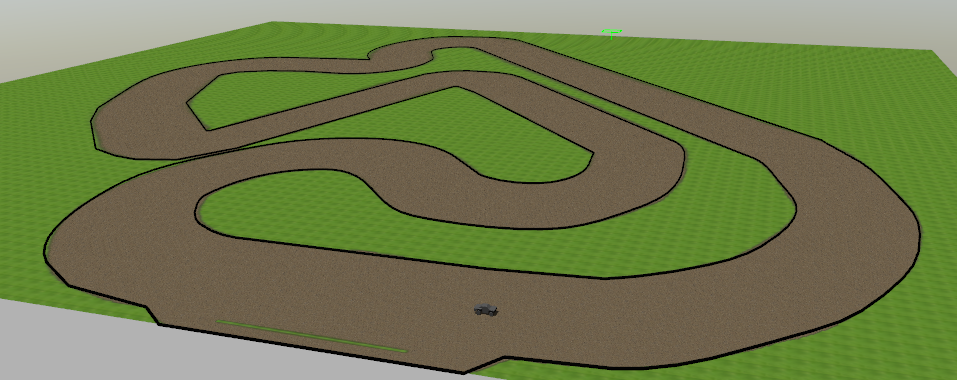}}
    % \vspace{1mm}
    % \subfloat[Track F \label{fig:RSS_map}]{\includegraphics[width=0.17\textwidth]{figs/RSS_map.png}}
    \caption{\textit{a)} The $1/5$ scaled ground vehicle used for experiments and \textit{b)} the track used for training (Track A) and \textit{c-g)} tracks used for testing (Track B, C, D, and E) %, and F).
    Track D and E %, and F 
    are from the ROS Gazebo simulator. Note that Track E is a simulated version of Track A, and Track D is a simulated version of Track B.}\label{fig:tracks}
\end{figure}

For a fair comparison, we trained all models with the same dataset used in \cite{Drews2019RAL}. 
The 90k data set consists of a vehicle running MPPI around a 170m-long track, the Track A.
It includes various lighting conditions, and views on the track.
Also shown in the supplementary video, the testing environment includes different shadow conditions and all the ruts, rocks, leaves, and grass on the dirt track provide various textures.
With a learning rate of 0.001, Adam \cite{adam} was used as an optimizer for training.
All the models converged with a training loss smaller than $5e^{-3}$ after $400$ epochs.

%%%%%%%%% Description of training data %%%%%%%%%%
% Our approach is more data-efficient to perform learning-based autonomous driving task compared to \cite{Drews2017corl, Drews2019RAL, Pan2018RSS}. 
% \cite{Drews2017corl, Pan2018RSS} required 300k and 12k of an autonomous driving dataset to train their model to perform their task in a 70m-long oval-shaped track (Track B). \cite{Drews2019RAL} trained their model with 90k of the dataset to run in a 170m-long track (Track A).
% For the real-world experiments, we trained our network with the same 90k data \cite{Drews2019RAL} used. The high-speed autonomous driving data running MPPI was collected at Track A \Cref{fig:CCRF}. We trained the same model \cite{Pan2018RSS} but with slightly different input image size.

% All of the previous learning-based autonomous driving work \cite{Drews2017corl, Drews2019RAL, Pan2018RSS} did not show the true generalizability of their methods. \cite{Drews2017corl, Drews2019RAL} collected almost every possible scenario in their large scale training dataset, including 64 different runs spanning 9 different days over the course of 8 months. \cite{Pan2018RSS} trained their end-to-end controller model with an online fashion so that the training data could possibly explore the state space as much as possible. However, these models were trained on the same track and tested on the same track.

The rest of this section will explore how each method performed on each track. These methods were compared over various real and simulated datasets including the TORCS open source driving simulator \cite{torcs} dataset and the KITTI dataset \cite{kitti}. 
For the TORCS dataset, we used the baseline test set collected by \cite{torcsdataset}.
All hardware experiments were conducted using our 1/5 scale autonomous vehicle test platform (\Cref{fig:car}) \cite{Autorally}.

\subsection{Costmap Prediction}

\begin{figure}
    \centering
    \vspace{-0.1cm}
    \subfloat{\includegraphics[width=0.09\textwidth]{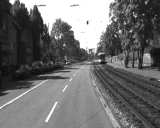}}
    \hspace{0.1mm}
    \subfloat{\includegraphics[width=0.09\textwidth]{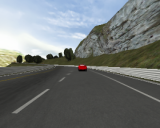}}
    \hspace{0.1mm}
    \subfloat{\includegraphics[width=0.09\textwidth]{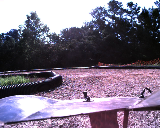}}
    \hspace{0.1mm}
    \subfloat{\includegraphics[width=0.09\textwidth]{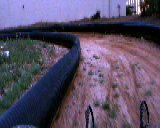}}
    \hspace{0.1mm}
    \subfloat{\includegraphics[width=0.09\textwidth]{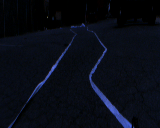}}
    \\
    \subfloat{\includegraphics[width=0.09\textwidth]{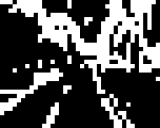}}
    \hspace{0.1mm}
    \subfloat{\includegraphics[width=0.09\textwidth]{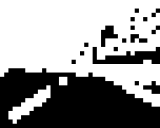}}
    \hspace{0.1mm}
    \subfloat{\includegraphics[width=0.09\textwidth]{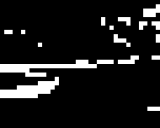}}
    \hspace{0.1mm}
    \subfloat{\includegraphics[width=0.09\textwidth]{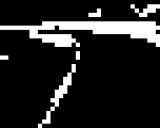}}
    \hspace{0.1mm}
    \subfloat{\includegraphics[width=0.09\textwidth]{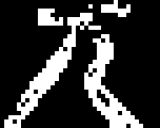}}
    \\
    \subfloat{\includegraphics[width=0.09\textwidth]{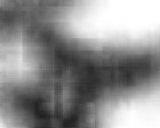}}
    \hspace{0.1mm}
    \subfloat{\includegraphics[width=0.09\textwidth]{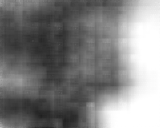}}
    \hspace{0.1mm}
    \subfloat{\includegraphics[width=0.09\textwidth]{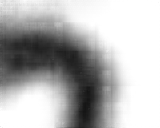}}
    \hspace{0.1mm}
    \subfloat{\includegraphics[width=0.09\textwidth]{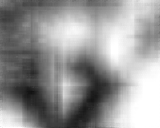}}
    \hspace{0.1mm}
    \subfloat{\includegraphics[width=0.09\textwidth]{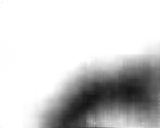}}
    \\
    \caption{\textit{Top:} From \textit{Left}: KITTI, TORCS, Track A, Track B, and Track C. \textit{Middle}: \ac{AIRL} (ours) costmap generation. \textit{Bottom}: Direct costmap prediction from \ac{ACP} \cite{drews2019visual}. \ac{E2E}\ac{IL} does not provide a costmap output so it is not pictured here. Note that \ac{ACP} works well on Track A data, which it was trained on.}
    \vspace{-0.3cm}
    \label{fig:costmap_comparison}
\end{figure}

We compare the methods mentioned in \Cref{sec:methods} on the following scenarios:
% \begin{inparaenum}[i)]
\begin{enumerate}
    \item Track A (real world complex track)
    \item Track B (real world oval track)
    \item Track C (real world pavement)
    \item Track D (ROS gazebo, simulated Track B)
    \item Track E (ROS gazebo, simulated Track A)
    % \item Track F (gazebo, simulated)
    \item TORCS driving simulator \cite{torcs}
    \item KITTI dataset \cite{kitti}.
\end{enumerate}
% \end{inparaenum}
The pictures of each track can be found in \Cref{fig:tracks}.

We first ran our proposed costmap model \ac{AIRL} and the benchmark model \ac{ACP} \cite{drews2019visual} on various datasets to show reasonable outputs in varied environments. The datasets used are KITTI, TORCs, Track A, Track B, and Track C as shown in \Cref{fig:costmap_comparison}. \ac{AIRL} produced costmaps that are interpretable by humans; \textit{\textbf{Assumption 1}} holds.
The predicted costmap of the \ac{ACP} is interpreted similarly to our method. The vehicle is located at the bottom middle of the costmap and black represents the low-cost region, white represents the high-cost. The difference is that \ac{ACP} produces a top-down-view/bird-eye-view costmap, whereas our method, \ac{AIRL}, produces a driver-view costmap.
\ac{ACP} produced clear cost maps models in Track A (which it was trained on) and Track C, though Track C's costmap was incorrect. These results show an inability for \ac{ACP} to generalize to varied different environments whereas our method produces similar looking costmaps throughout.

\subsection{Autonomous Driving}
\label{subsection:AutonomousDriving}

\begin{figure}
    \centering
    \includegraphics[width=0.3\textwidth]{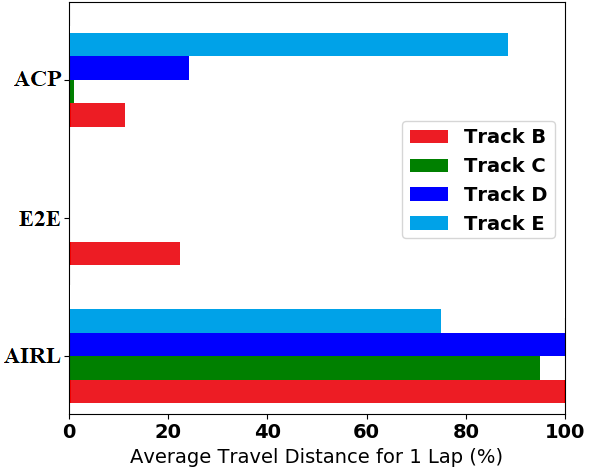}
    \vspace{-0.3cm}
    \caption{Generalization results of autonomous driving on tracks outside of the training dataset. AIRL (ours) outperforms the other two state-of-the-arts in most cases.}
    \vspace{-0.3cm}
    \label{fig:total_results}
\end{figure}

We then took all three methods and drove them on Tracks B, C, D, and E. %, and F.
Since it is too obvious and already reported in \cite{drews2019visual, Pan2018RSS}, here we do not report how well each method does on training data (Track A).
For Tracks B, D, and E, we ran each algorithm in both clockwise and counter-clockwise for 20 lap attempts and measured the average travel distance. We can see in \Cref{fig:total_results}, that our approach was the only method that was able to finish the whole lap of driving Track B and D. %, and F.
Compared to other methods, our proposed method, \ac{AIRL}, tended to hug track boundaries closely, presumably because of the sparsity of our costmaps.
% We note in Appendix \Cref{subsec:fail} a failure state of our method and potential reasons for it.

The parameters used for AIRL's MPPI in image space for all trials are as follows and they correspond to the original MPPI paper \cite{mppi17}: 
$\mathbf{K}=2, v_{x}^d=5.0m/s$ for off-road driving, 
$v_{x}^d=2.5m/s$ for on-road driving, 
$\Delta_t=0.02, \Tau = 60, \Sigma_{steer}=0.3, \Sigma_{throttle}=0.35, C_{speed}=1.8$, and $C_{crash}=0.9$. $\Tau$ was set to correlate to approximately $6m$ long trajectories, as this covers almost all the drivable area in the camera view (see \Cref{fig:mppi_imagespace}).

The benchmark method \ac{ACP} failed to drive more than half of Track B because the predicted costmap was not stable as seen in \Cref{fig:costmap_comparison} Track B results.
% and \Cref{fig:paul_good_bad}.
When looking at the costmaps generated from \ac{ACP} in \Cref{fig:costmap_comparison}, we can see that the model trained on Track A is not generating a clear costmap when tested on Track B.

% \begin{figure}[H]
%     \centering
%     \includegraphics[width=0.12\textwidth]{figs/Paul_input_img_000675.png}
%     \hspace{0.5mm}
%     \includegraphics[width=0.12\textwidth]{figs/Paul_marietta_output000675.png}
%     \hspace{0.5mm}
%     \includegraphics[width=0.12\textwidth]{figs/Paul_ccrf_output000675.png}
%     \vspace{-0.25cm}
%     \caption{Costmap prediction differences of \ac{ACP} \cite{drews2019visual} due to different training data. \textit{Left}: Resized and cropped input image from Track B (test data). \textit{Middle}:  costmap prediction from \ac{ACP} trained on Track B. \textit{Right}: costmap prediction from \ac{ACP} trained on Track A. }
%     \label{fig:paul_good_bad}
% \end{figure}

We ensured that the poor performance of the benchmark model \ac{ACP} was not due to improper tuning of \ac{MPPI} by training another model of \cite{drews2019visual} on Track B data only.
We then tuned \ac{MPPI} with this model and drove it around Track B successfully for 10 laps straight before being manually stopped.
After this verification of \ac{MPPI} parameters, we applied the same parameters to \ac{ACP}. 
Unfortunately, we did not see the same track coverage with properly tuned \ac{MPPI}.

Overall, \ac{ACP} performed best on Track E, which is a simulated version of the track it was trained upon, Track A. The other Tracks have a similar issue to Track B, i.e. unclear costmaps. 

% With the model trained on the same track and tested on the same one, MPPI was able to finish the whole lap without any collision because the predicted costmap was very stable.
% With this working-MPPI, we used the same MPPI parameters but only changed the model to one which was trained only with the Track A data. %\Cref{fig:CCRF} data.

Surprisingly, the network trained through end-to-end imitation learning  (\ac{E2E}\ac{IL}) was able to drive up to half of a lap on Track B, but in all of the sim datasets (Tracks D and E), it did not move. This is most likely due to images not matching the training distribution of images. The ability to drive on Track B is most likely due to the images being somewhat similar to Track A as can be seen in \Cref{fig:costmap_comparison}.

We also verified the generalization of each method at a totally new on-road environment, Track C. We made a 30m-long zigzag lane on the tarmac to look like a real road situation. Since the training data was collected at Track A, an off-road dirt track, the tarmac surface is totally new; in addition, the boundaries of the course changed from black plastic tubes to taped white lanes. The width between the boundaries varied from 0.5 m to 1.5 m and was in general much tighter than the off-road tracks. Moreover, we ran our algorithm in the late afternoon, which has very different lighting conditions compared to the training data as seen in \Cref{fig:costmap_comparison}.
Even under this large change of environments, \ac{AIRL} with \ac{MPPI} was still able to accomplish a lane-keeping task most of the time, whereas the other two methods, \ac{E2E}\ac{IL} and \ac{ACP} immediately failed the task.

\section{DISCUSSION}
\label{sec:discussion}
We analyze that the false prediction of the benchmarks \cite{Pan2018RSS} and \cite{drews2019visual} comes from the covariate shift of the input image distribution.
% Why did other methods fail? If we split the typical autonomy pipeline in two, we can split it into \textit{a)} a pipeline from sensor measurements to task-specific objective functions generation, and \textit{b)} a pipeline from objective functions to corresponding optimal path and control.
% In the costmap learning approach \ac{ACP}, \cite{drews2019visual} uses deep learning to replace the pipeline \textit{a)}, and uses an MPC controller to handle \textit{b)}.
% However, if the first pipeline fails to produce a correct objective function, the second part of path planning will calculate a wrong result and fail the task, no matter how well the controller or path planner is tuned.
% In our experiments, we saw this happens frequently (\Cref{fig:paul_good_bad}) and we analyze that this false prediction comes from the covariate shift of the input image distribution.
% The input might be new to the network, i.e. the training data did not include that specific image, or the trained network did not correctly learn the mapping from those input data to a corresponding output.
% Our approach also replaces the first pipeline \textit{a)} with deep learning, but it always outputs a correct costmap through our proposed method, \ac{AIRL}, no matter whether the input is corrupted or new. 
In the network trained through end-to-end imitation learning (\ac{E2E}\ac{IL}), although the middle layer outputs meaningful features/heatmap, a small change of each middle layer’s activation coming from a novel input results in a random or false NN output.
For this reason, we cannot use the whole (same) NN parameters used in the \ac{E2E}\ac{IL} training phase. However, without throwing it away, we can still use the CNN portion of the original network for feature extraction.
We saw in our experiments that the values of relevance scores of each pixel in the \textit{forward heatmap} changes when it sees an out-of-sample input. This results in a failure of the output prediction of the whole \ac{DNN} model.
However, although the CNN parameters did not learn how much activation each neuron needs for data outside the training set, it surely learned the task-related feature extractions; the feature extractor part is still functional.
Although the relevance `scores' slightly changed, the activated `features' are still unaffected.
This is why \ac{AIRL} shows great generalizability of producing a costmap. With a binary filter, it performs even better producing more distinguishable and robust costmaps.

\section{CONCLUSION}
\label{sec:conclusion}
We introduced an \acl{AIRL} framework using deep \acp{CNN}.
By utilizing the powerful feature extracting nature of CNNs, our method automates the generalizable cost function learning through Imitation Learning by linking the extracted features to the cost function used in optimal control.
Transferring pixel heatmaps of middle layers in a IL-trained network as a cost function holds the promises to automate the feature extraction and cost function design for other vision-based tasks as well.
Beyond autonomous driving, any vision-based MDP problems, e.g. drones, legged robots, and manipulators with various tasks, are possible applications of the proposed approach.

\addtolength{\textheight}{-7cm}
\bibliographystyle{IEEEtran}
\bibliography{airl}  % .bib

\begin{thebibliography}{10}
\providecommand{\url}[1]{#1}
\csname url@samestyle\endcsname
\providecommand{\newblock}{\relax}
\providecommand{\bibinfo}[2]{#2}
\providecommand{\BIBentrySTDinterwordspacing}{\spaceskip=0pt\relax}
\providecommand{\BIBentryALTinterwordstretchfactor}{4}
\providecommand{\BIBentryALTinterwordspacing}{\spaceskip=\fontdimen2\font plus
\BIBentryALTinterwordstretchfactor\fontdimen3\font minus
  \fontdimen4\font\relax}
\providecommand{\BIBforeignlanguage}[2]{{%
\expandafter\ifx\csname l@#1\endcsname\relax
\typeout{** WARNING: IEEEtran.bst: No hyphenation pattern has been}%
\typeout{** loaded for the language `#1'. Using the pattern for}%
\typeout{** the default language instead.}%
\else
\language=\csname l@#1\endcsname
\fi
#2}}
\providecommand{\BIBdecl}{\relax}
\BIBdecl

\bibitem{Pan2018RSS}
Y.~Pan, C.-A. Cheng, K.~Saigol, K.~Lee, X.~Yan, E.~A. Theodorou, and B.~Boots,
  ``{Agile Autonomous Driving using End-to-End Deep Imitation Learning},''
  \emph{Robotics: Science and Systems}, 2018.

\bibitem{drews2019visual}
P.~Drews, ``{Visual Attention for High Speed Driving},'' Ph.D. dissertation,
  Georgia Institute of Technology, 2019.

\bibitem{Drews2017corl}
P.~Drews, G.~Williams, B.~Goldfain, E.~A. Theodorou, and J.~M. Rehg,
  ``{Aggressive Deep Driving: Combining Convolutional Neural Networks and Model
  Predictive Control},'' in \emph{1st Annual Conference on Robot Learning, CoRL
  2017, Mountain View, California, USA, November 13-15, 2017, Proceedings},
  2017, pp. 133--142.

\bibitem{Drews2019RAL}
P.~{Drews}, G.~{Williams}, B.~{Goldfain}, E.~A. {Theodorou}, and J.~M. {Rehg},
  ``{Vision-Based High-Speed Driving With a Deep Dynamic Observer},''
  \emph{IEEE Robotics and Automation Letters}, vol.~4, no.~2, pp. 1564--1571,
  04 2019.

\bibitem{LevineVisuomotor}
S.~Levine, C.~Finn, T.~Darrell, and P.~Abbeel, ``{End-to-End Training of Deep
  Visuomotor Policies},'' \emph{Journal of Machine Learning Research}, vol.~17,
  no.~39, pp. 1--40, 2016.

\bibitem{giusti2016machine}
A.~Giusti, J.~Guzzi, D.~Ciresan, F.-L. He, J.~P. Rodriguez, F.~Fontana,
  M.~Faessler, C.~Forster, J.~Schmidhuber, G.~Di~Caro, D.~Scaramuzza, and
  L.~Gambardella, ``{A Machine Learning Approach to Visual Perception of Forest
  Trails for Mobile Robots},'' \emph{IEEE Robotics and Automation Letters},
  2016.

\bibitem{lee2020pixelmpc}
K.~{Lee}, J.~{Gibson}, and E.~A. {Theodorou}, ``{Aggressive Perception-Aware
  Navigation using Deep Optical Flow Dynamics and PixelMPC},'' \emph{IEEE
  Robotics and Automation Letters}, 2020.

\bibitem{bojarski2016end}
M.~Bojarski, D.~{Del Testa}, D.~Dworakowski, B.~Firner, B.~Flepp, P.~Goyal,
  L.~D. Jackel, M.~Monfort, U.~Muller, J.~Zhang, X.~Zhang, J.~Zhao, and
  K.~Zieba, ``{End to End Learning for Self-Driving Cars},'' \emph{arXiv}, 04
  2016.

\bibitem{xu2017end}
H.~Xu, Y.~Gao, F.~Yu, and T.~Darrell, ``{End-to-end learning of driving models
  from large-scale video datasets},'' in \emph{Proceedings of the IEEE
  conference on computer vision and pattern recognition}, 2017, pp. 2174--2182.

\bibitem{Ollis2008outdoor}
M.~{Ollis}, W.~H. {Huang}, M.~{Happold}, and B.~A. {Stancil}, ``{Image-based
  path planning for outdoor mobile robots},'' in \emph{2008 IEEE International
  Conference on Robotics and Automation}, 05 2008, pp. 2723--2728.

\bibitem{arora2018survey}
S.~Arora and P.~Doshi, ``{A survey of inverse reinforcement learning:
  Challenges, methods and progress},'' \emph{arXiv preprint arXiv:1806.06877},
  2018.

\bibitem{Babes2011IRL}
{M. Babes, V. Marivate, M. Littman, and K. Subramanian.}, ``{Apprenticeship
  learning about multiple intentions.}'' in \emph{Proceedings of the 28th
  International Conference on Machine Learning}, 2011.

\bibitem{Kuderer2015ICRA}
M.~{Kuderer}, S.~{Gulati}, and W.~{Burgard}, ``{Learning driving styles for
  autonomous vehicles from demonstration},'' in \emph{2015 IEEE International
  Conference on Robotics and Automation (ICRA)}, 05 2015, pp. 2641--2646.

\bibitem{DAgger}
S.~Ross, G.~J. Gordon, and J.~A. Bagnell, ``{A Reduction of Imitation Learning
  and Structured Prediction to No-Regret Online Learning},'' in
  \emph{Proceedings of the 14th International Conference on Artificial
  Intelligence and Statistics}, ser. JMLR, vol.~15, Fort Lauderdale, FL, USA,
  2011.

\bibitem{lee2019earlyfailure}
K.~{Lee}, K.~{Saigol}, and E.~A. {Theodorou}, ``Early failure detection of deep
  end-to-end control policy by reinforcement learning,'' in \emph{2019
  International Conference on Robotics and Automation (ICRA)}, 05 2019, pp.
  8543--8549.

\bibitem{lee2019ensemble}
K.~Lee, Z.~Wang, B.~I. Vlahov, H.~K. Brar, and E.~A. Theodorou, ``Ensemble
  bayesian decision making with redundant deep perceptual control policies,''
  \emph{18th IEEE International Conference on Machine Learning and Applications
  (ICMLA)}, 2019.

\bibitem{lee2019papc}
K.~Lee, G.~N. An, V.~Zakharov, and E.~A. Theodorou, ``Perceptual
  attention-based predictive control,'' \emph{3rd Conference on Robot Learning
  (CoRL)}, 2019.

\bibitem{mppi17}
G.~{Williams}, N.~{Wagener}, B.~{Goldfain}, P.~{Drews}, J.~M. {Rehg},
  B.~{Boots}, and E.~A. {Theodorou}, ``{Information theoretic MPC for
  model-based reinforcement learning},'' in \emph{2017 IEEE International
  Conference on Robotics and Automation (ICRA)}, 05 2017, pp. 1714--1721.

\bibitem{mppi16}
G.~Williams, P.~Drews, B.~Goldfain, J.~M. Rehg, and E.~A. Theodorou,
  ``{Aggressive driving with model predictive path integral control},''
  \emph{2016 IEEE International Conference on Robotics and Automation (ICRA)},
  2016.

\bibitem{coordtransformbook}
E.~Trucco and A.~Verri, \emph{Introductory Techniques for 3-D Computer
  Vision}.\hskip 1em plus 0.5em minus 0.4em\relax Upper Saddle River, NJ, USA:
  Prentice Hall PTR, 1998.

\bibitem{interpretingDNN}
G.~Montavon, W.~Samek, and K.-R. Müller, ``{Methods for interpreting and
  understanding deep neural networks},'' \emph{Digital Signal Processing},
  vol.~73, pp. 1 -- 15, 2018.

\bibitem{explainableAI}
W.~Samek, G.~Montavon, A.~Vedaldi, L.~K. Hansen, and K.-R. M{\"u}ller,
  \emph{{Explainable AI: Interpreting, Explaining and Visualizing Deep
  Learning}}.\hskip 1em plus 0.5em minus 0.4em\relax Springer, Cham, 2019, vol.
  11700.

\bibitem{samek2017evalvis}
W.~{Samek}, A.~{Binder}, G.~{Montavon}, S.~{Lapuschkin}, and K.~{Müller},
  ``{Evaluating the Visualization of What a Deep Neural Network Has Learned},''
  \emph{IEEE Transactions on Neural Networks and Learning Systems}, vol.~28,
  no.~11, pp. 2660--2673, 2017.

\bibitem{lrp}
S.~Bach, A.~Binder, G.~Montavon, F.~Klauschen, K.-R. Müller, and W.~Samek,
  ``{On Pixel-Wise Explanations for Non-Linear Classifier Decisions by
  Layer-Wise Relevance Propagation},'' \emph{PLOS ONE}, vol.~10, no.~7, 2015.

\bibitem{adam}
D.~P. Kingma and J.~Ba, ``{Adam: {A} Method for Stochastic Optimization},''
  \emph{Proceedings of the 3rd International Conference on Learning
  Representations (ICLR)}, vol. abs/1412.6980, 2014.

\bibitem{torcs}
B.~Wymann, C.~Dimitrakakisy, A.~Sumnery, and C.~Guionneauz, ``{TORCS: The open
  racing car simulator},'' 2015.

\bibitem{kitti}
A.~Geiger, P.~Lenz, C.~Stiller, and R.~Urtasun, ``{Vision meets Robotics: The
  KITTI Dataset},'' \emph{International Journal of Robotics Research (IJRR)},
  2013.

\bibitem{torcsdataset}
C.~Chen, A.~Seff, A.~Kornhauser, and J.~Xiao, ``{DeepDriving: Learning
  Affordance for Direct Perception in Autonomous Driving},'' \emph{Proceedings
  of 15th IEEE International Conference on Computer Vision}, 2015.

\bibitem{Autorally}
B.~Goldfain, P.~Drews, C.~You, M.~Barulic, O.~Velev, P.~Tsiotras, and J.~M.
  Rehg, ``{AutoRally: An Open Platform for Aggressive Autonomous Driving},''
  \emph{IEEE Control Systems Magazine}, vol.~39, no.~1, pp. 26--55, 02 2019.

\end{thebibliography}

\onecolumn
\section*{Citations}
\hspace{-0.3cm}Plain Text:
\\ \\
K. Lee, B. Vlahov, J. Gibson, J. M. Rehg, and E. A. Theodorou, ``Approximate Inverse Reinforcement Learning from Vision-based Imitation Learning,'' The 2021 International Conference on Robotics and Automation.
\\ \\
BibTeX:
\\ \\
@ARTICLE$\{$lee2021approximate,
\\author=$\{$Keuntaek $\{$Lee$\}$ and Bogdan $\{$Vlahov$\}$ and Jason $\{$Gibson$\}$ and James M. $\{$Rehg$\}$ and Evangelos A. $\{$Theodorou$\}$$\}$,
\\journal=$\{$The 2021 International Conference on Robotics and Automation$\}$,
\\title=$\{$$\{$Approximate Inverse Reinforcement Learning from Vision-based Imitation Learning$\}$$\}$,
\\year=$\{$2021$\}$
\\$\}$

\end{document}